\documentclass[10pt,twocolumn,letterpaper]{article}

\usepackage[pagenumbers]{iccv} 
\usepackage{multirow}
\usepackage{colortbl}
\usepackage{comment}

\newcommand{\datasetname}{AnimalClue\xspace}

\makeatletter
\renewcommand{\paragraph}{%
  \@startsection{paragraph}{4}%
  {\z@}{0.25em}{-1em}%
  {\normalfont\normalsize\bfseries}%
}
\makeatother

\definecolor{iccvblue}{rgb}{0.21,0.49,0.74}
\usepackage[pagebackref,breaklinks,colorlinks,allcolors=iccvblue]{hyperref}
\usepackage{cuted}
\def\paperID{668} 
\def\confName{ICCV}
\def\confYear{2025}

\title{\datasetname: Recognizing Animals by their Traces}

\author{\large Risa Shinoda\textsuperscript{\rm 1,2,4},
Nakamasa Inoue\textsuperscript{\rm 3,4},
Iro Laina\textsuperscript{\rm 5},
Christian Rupprecht\textsuperscript{\rm 5},
Hirokatsu Kataoka\textsuperscript{\rm 4,5}
\\
\textsuperscript{\rm 1}The University of Osaka
\textsuperscript{\rm 2}Kyoto University
\textsuperscript{\rm 3}Tokyo Institute of Technology\\
\textsuperscript{\rm 4}National Institute of Advanced Industrial Science and Technology (AIST)\\
\textsuperscript{\rm 5}Visual Geometry Group, University of Oxford
}

\begin{document}
\maketitle
\begin{strip}\centering
\vspace{-0.6in}
\includegraphics[width=\textwidth]{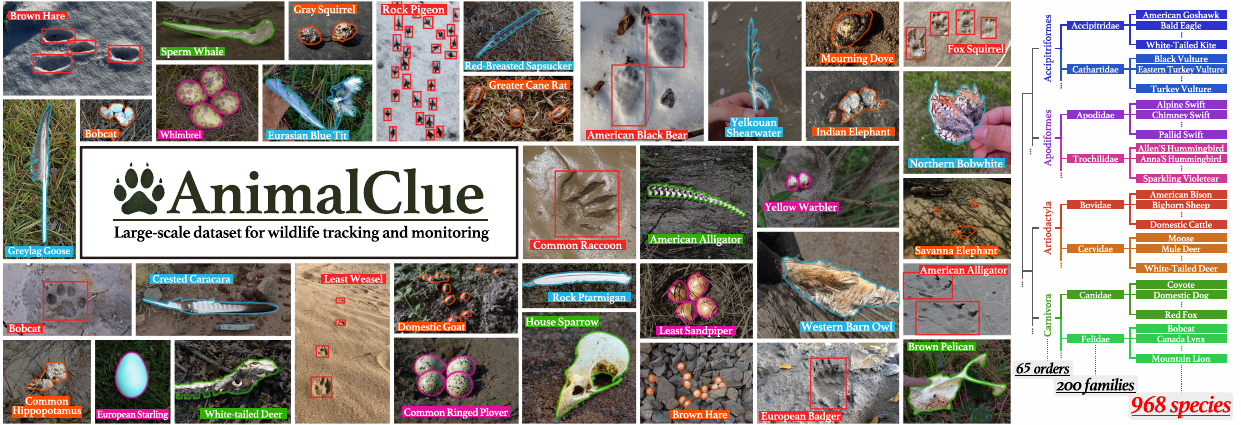}
 \vspace{-.2in}
\captionof{figure}{\textbf{Example images of \datasetname.} We present \datasetname, a dataset designed for identifying animal species based on their traces. Our dataset includes footprints, feces, eggs, bones, and feathers, totaling 159,605 bounding boxes from 968 animal species. We also annotate 22 traits such as habitat, diet, and activity pattern.
We establish four benchmarks for evaluating models: classification, detection, instance segmentation, and traits classification. \datasetname~has the potential to advance research in animal tracking, an area that remains underexplored due to the limited availability of publicly accessible datasets.
}
\end{strip}
\begin{abstract}
Wildlife observation plays an important role in biodiversity conservation, necessitating robust methodologies for monitoring wildlife populations and interspecies interactions. Recent advances in computer vision have significantly contributed to automating fundamental wildlife observation tasks, such as animal detection and species identification. However, accurately identifying species from indirect evidence like footprints and feces remains relatively underexplored, despite its importance in contributing to wildlife monitoring. To bridge this gap, we introduce \textbf{AnimalClue}, the first large-scale dataset for species identification from images of indirect evidence. Our dataset consists of 159,605 bounding boxes encompassing five categories of indirect clues: footprints, feces, eggs, bones, and feathers. It covers 968 species, 200 families, and 65 orders. Each image is annotated with species-level labels, bounding boxes or segmentation masks, and fine-grained trait information, including activity patterns and habitat preferences. Unlike existing datasets primarily focused on direct visual features (e.g., animal appearances), AnimalClue presents unique challenges for classification, detection, and instance segmentation tasks due to the need for recognizing more detailed and subtle visual features. In our experiments, we extensively evaluate representative vision models and identify key challenges in animal identification from their traces. Our dataset and code are available at \url{https://dahlian00.github.io/AnimalCluePage/}
\end{abstract}    

\section{Introduction}
\label{sec:intro}
Conserving wildlife is essential for sustaining ecological balance and maintaining healthy ecosystems. However, numerous human activities, such as habitat destruction, chemical pollution, and the depletion of food resources, pose severe threats to wildlife species. To mitigate these impacts, computer vision technologies are increasingly expected to play a significant role by automating comprehensive and continuous observation of wildlife populations. For example, animal detection models implemented with unobtrusive cameras can remotely monitor animals without disturbing their natural behaviors. This enables researchers to detect changes in behavioral patterns.

Nevertheless, identifying a wide range of animal species remains challenging, as most models overlook nocturnal animals or exhibit camouflage behaviors. Wildlife monitoring can leverage indirect biological evidence, such as footprints, bone fragments, and feathers to address this challenge. This approach is widely employed in ecological surveys; however, its scalability is limited due to heavy reliance on manual human inspection. Consequently, computer vision models capable of accurately identifying animal species from such indirect evidence are in high demand, as they could significantly expand the scale and effectiveness of wildlife monitoring efforts.

Comprehensive datasets have contributed to the progress of animal identification using computer vision~\cite{8579012,shinoda2024petface,KhoslaYaoJayadevaprakashFeiFei_FGVC2011,parkhi12a,WahCUB_200_2011,7298658,Chen_2023_CVPR,7552915,Ng_2022_CVPR}.
However, the majority of computer vision datasets for animal studies have focused on direct visual observation tasks such as animal identifications~\cite{AerialCattle2017, ATRW,fishnet}, activity recognition~\cite{Chen_2023_CVPR,ani11020485}, image-text alignment~\cite{INQUIRE,gabeff2023wildclip}, and pose estimation~\cite{animalpose,yang2022apt,yu2021ap,ye2023superanimal}. 
Several datasets involve indirect animal clues, but their focus is primarily limited to classification tasks with a small number of species~\cite{AutomatedEggClassification, 10.1145/3581783.3612708,ani13101660,AutomatedEggClassification}. Given that classification, detection, and segmentation are established foundational tasks in the field of computer vision, developing a new indirect-evidence dataset that incorporates these tasks across a diverse range of species has significant potential to advance research.

Motivated by this, we introduce \textbf{AnimalClue}, a dataset that includes five major types of animal traces: footprints, feces, eggs, bones, and feathers, with sophisticated bounding boxes and segmentation masks.
It encompasses 968 species with a total of 159,605 bounding boxes and 141,314 annotation masks.
Furthermore, we annotate 22 species-specific traits, including habitat, activity pattern, and diet, to facilitate the research for animal tracking. Analyzing animal traits depending on their animal traces has not been previously explored.

We establish four benchmarks: classification, detection, instance segmentation, and habitat prediction for animal traces.
Our findings indicate that models trained on our dataset develop the ability to recognize animal traces, which differ significantly from the visual features of the animals themselves. However, our dataset remains challenging:
(a) Learning rare species is challenging, making generalization to rare species difficult.
(b) Detection and instance segmentation tasks are particularly challenging, with models struggling to achieve high mAP.
The highest mAP for order detection is 0.57, and the best order instance segmentation mAP is 0.48.
We will make the dataset publicly available for research purposes.
\section{Related Work}

\begin{table*}[t]
    \centering 
    \begin{tabular}{lccccc} 
    \toprule
    {Dataset} & {Track Type} & {\#Species} & {\#Bbox} &Task&{\#Traits}  \\
      \midrule
OpenAnimalTracks~\cite{shinoda2024openanimaltracks}&Footprint&18&3,579&CLS, DET&0\\

    DFML~\cite{ani13101660}&Feces&1&1,623&CLS&1\\
  
      Automated Egg Classification~\cite{AutomatedEggClassification}&Egg&2&2,943&CLS&0\\

       Skull2Animal dataset~\cite{10.1145/3581783.3612708} &Bone&4&4,962&CLS&0\\
FeathersV1~\cite{belko2020feathersdatasetfinegrainedvisual}&Feather&595& 28,272&CLS&0\\
            \rowcolor[gray]{0.8} \datasetname~(Ours) & All 5 Types&968&159,605&CLS, DET, SEG&22\\
      \bottomrule
      \end{tabular}

    \caption{\textbf{Comparison with previous animal tracking datasets.} CLS, DET, and SEG indicate classification, detection, and instance segmentation, respectively. Our \datasetname contains diverse species and a larger number of bounding boxes.}
  \label{tb:related_dataset}
  \vspace{-10pt}
\end{table*}
\label{sec:related}
\paragraph{Animal species identification.}
Automating the identification of animal species is important for effective wildlife monitoring. Species labels can be obtained from various sources, including images~\cite{8579012,shinoda2024petface,KhoslaYaoJayadevaprakashFeiFei_FGVC2011,parkhi12a,WahCUB_200_2011,7298658}, videos~\cite{Chen_2023_CVPR,7552915,Ng_2022_CVPR}, 3D models~\cite{Xu_2023_ICCV}, and pose annotations~\cite{Cao_2019_ICCV,yu2021ap}. 
These datasets and benchmarks contribute to developing models and applying computer vision techniques in the animal domain. While direct animal identification has been well-explored, there remains room to investigate methods that identify animals indirectly, such as through animal traces left behind.
In the rest of this section, we discuss related work on the identification of these traces.

\paragraph{Footprint identification.}
Recognizing animals from their footprints has been the subject of previous research, as humans also leave traces of their footprints, which often need to be identified~\cite{9421542,8609926}.
Similarly, identifying animal footprints is crucial for wildlife monitoring and conservation efforts.
Previous works attempt individual identification~\cite{whiterhino,Jewell2016SpottingCI}, species identification~\cite{whiterhino}, and sex identification~\cite{panda}.
While previous research has explored the utility of footprints as animal clues, the associated images are not openly available. Recently, OpenAnimalTracks~\cite{shinoda2024openanimaltracks} was introduced as a classification and detection dataset for animal footprints. However, our \datasetname~differs from OpenAnimalTracks in several key aspects: (i) species distribution: \datasetname~includes 117 species compared to OpenAnimalTracks' 18 species; (ii)  \datasetname~contains approximately five times more footprint bounding boxes, and additionally includes segmentation masks; and (iii) our dataset exclusively contains images licensed under Creative Commons.

\paragraph{Feces identification.}
Feces are a key indicator of health. In the medical field, several studies have explored automated feces analysis for human health~\cite{pmlr-v149-zhou21a,hachuel2019augmentinggastrointestinalhealthdeep,4649835}, focusing primarily on feces detection and health classification.
Recently, there has been research on automating the classification of chicken feces to assess health abnormalities into six categories~\cite{ani13193041}.
For dogs, the DFML dataset~\cite{ani13101660} comprises 1,623 photos of dog feces, yet the recognition of feces across multiple species remains relatively unexplored.
Considering the limited research efforts applying computer vision to animal feces compared to the human domain, \datasetname~presents a challenging dataset with the potential to advance animal feces recognition through computer vision.

\paragraph{Egg identification.}
Few previous studies have explored bird egg detection~\cite{egg_detection,cae_egg_type,Acoustic_egg}, and those that do typically focus on a single species due to their primary interest in agriculture rather than species identification for animal tracking. Moreover, these studies use data that are not publicly available.
As for openly available datasets, the Automated Egg Classification dataset~\cite{AutomatedEggClassification} includes 3k images of chicken and duck eggs. However, species identification from eggs remains largely unexplored, and our dataset aims to address this gap.

\paragraph{Bone identification.}
Bone recognition has gained significant attention in the medical field, particularly for human bones, with several datasets contributing to this area~\cite{Abedeen_2023,bone_age,rajpurkar2017mura}.
For animals, the Skull2Animal dataset~\cite{10.1145/3581783.3612708} includes skull images of four species (e.g., dogs, cats, leopards, and foxes), comprising 5k images. This dataset contributes to the prediction of skull images for animal species identification. For the purpose of animal identification, there are not enough publicly available datasets.

\paragraph{Feather identification.}
Feathers are key traces of birds, with their appearance varying depending on the bird species and the part of the body. During molting seasons, feathers can be found in abundance. FeathersV1~\cite{belko2020feathersdatasetfinegrainedvisual} is a dataset designed for fine-grained bird feather categorization, offering high-quality images to facilitate detailed classification. Our dataset differs from FeathersV1 in two significant ways: (i) most of our data is captured in the wild, not in controlled environments, making it more suitable and challenging for animal tracking, and (ii) we provide both bounding box and fine-grained pixel-level annotations.

\section{\datasetname~}
\label{sec:dataset}
In this section, we introduce \datasetname, for the purpose of wildlife conservation from multiple animal traces.
During dataset collection, we carefully sourced images from the Internet under the Creative Commons license.
For annotations, labels were verified by multiple citizen scientists, and we assigned precise bounding boxes and segmentation labels.

\subsection{Trace Types and Animal Species}
\label{sec:trace_type_and_animal_species}
\begin{figure*}
\centering
\includegraphics[width=\linewidth]{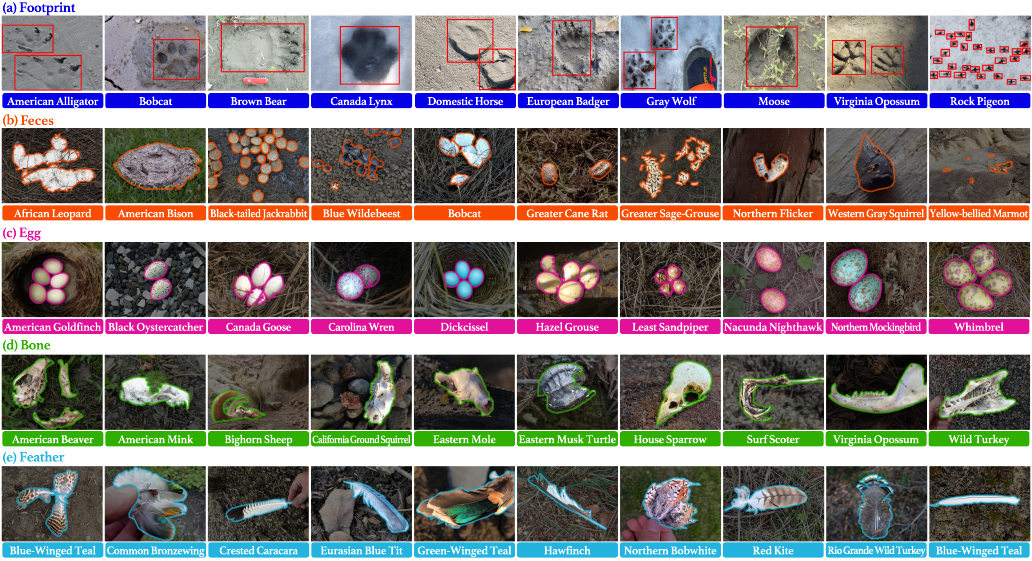}
\caption{\textbf{Example images and annotations on \datasetname.} The figures illustrate the five different animal clues, (a) footprints, (b) feces, (c) eggs, (d) bones, and (e) feathers, which are observed indirectly. There exist segmentation labels and bounding boxes for all clues except for footprints.
}
\label{fig:animal_images}
\vspace{-15pt}
\end{figure*}

In tracing the types of animal species, we have collected five clues from animals: footprints, feces, bones, eggs, and feathers. These types of animal clues are also commonly used to recognize animal species in wildlife conservation.
Figure~\ref{fig:animal_images} illustrates examples of the correspondence between animal traces and their respective species. For a complete species list, please refer to the supplementary material.

\subsection{Data Collection}
\label{sec:data_collection}
We have collected images from iNaturalist~\cite{iNaturalistWeb}, a web platform for citizen scientists focusing on wildlife. We chose iNaturalist for two key reasons:
i) Quality: The iNaturalist community comprises citizen scientists who ensure higher-quality data than random social media platforms. Images are reviewed by the iNaturalist community, ensuring reliable labels. We chose research-grade images, confirmed by other citizen scientists for the image and animal species pair.
ii) Licensing: Users can select their images' licenses.

We have selected several iNaturalist projects that focus on uploading images of animal traces. From these, we downloaded the images under the Creative Commons licenses, along with their corresponding animal species labels. 
We have removed any ambiguous or unclear images from \datasetname to improve data quality. Additionally, we exclude images containing human faces to protect privacy.
Please refer to the supplementary material for examples of removed images.
Importantly, we collect only research-grade images, where labels are verified by multiple researchers to ensure the quality of \datasetname.

\subsection{Data Annotation}
\label{sec:data_annotation}

\noindent\textbf{Annotation type.} 
Once we obtain the animal trace images and corresponding species labels, we proceed with bounding boxes and segmentation masks for object detection and instance segmentation tasks. Figure~\ref{fig:animal_images} illustrates the example images and their annotations on \datasetname. Different annotation types are applied flexibly depending on the type of animal trace. For animal footprints, we annotate each footprint with a bounding box. In contrast, we use segmentation masks for feces and feathers to capture more detailed features.
We limit footprint annotations to bounding boxes because footprints are traces rather than physical objects, and they are often ambiguous in some parts. In contrast, feces, bones, eggs, and feathers can be annotated with fine-grained, pixel-level segmentation masks. Defining clear bounding box groups for feces can be challenging, and in this sense, instance segmentation is helpful in determining boundaries.
\begin{figure*}[t]
  \centering
  \includegraphics[width=0.99\linewidth]{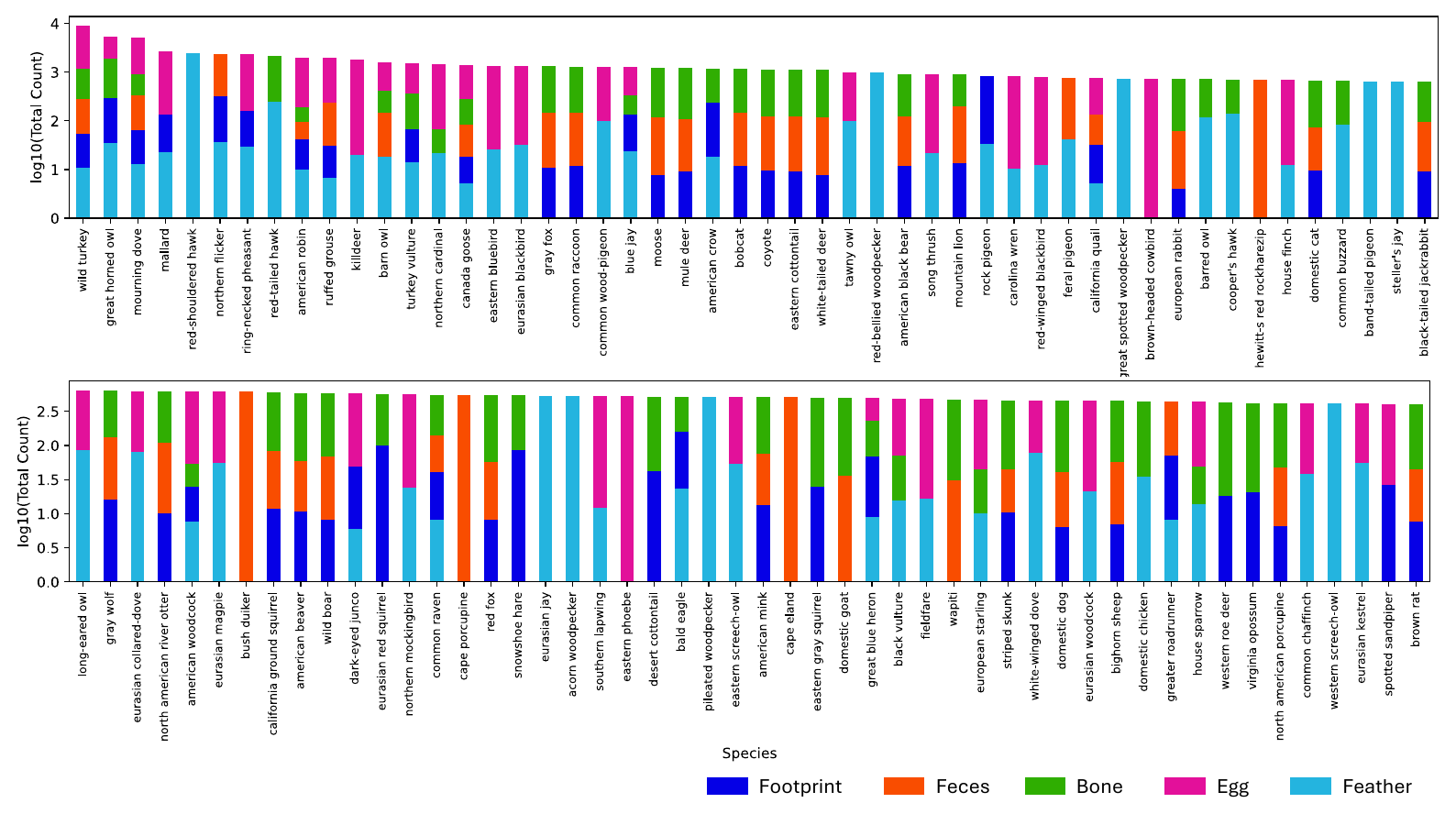}
\caption{\textbf{Species distribution in our \datasetname.} We illustrate the top 100 most frequent species, with the vertical axis representing log frequency.}
  \label{fig:fig2}
  \vspace{-15pt}
\end{figure*}

\noindent\textbf{Annotation process.}
For half of the footprint annotations, we use a third-party annotation service to assign bounding boxes, and authors cross-check all bounding boxes to ensure they are precise enough for model training. For the remaining footprint, feces, egg, bone, and feather annotations, the authors conducted the annotations.

For footprints, feces, and bones, we manually annotate from scratch. For eggs and feathers, we use the Segment Anything Model as an initial annotation, which is then reviewed by the authors. If an annotated mask is incorrect, we manually correct it.

\noindent\textbf{Fine-grained trait annotations.}
We annotated each species with multiple ecological and behavioral attributes to facilitate a comprehensive analysis of species traits. Specifically, we assigned the following attributes:
\begin{itemize}
\item Taxonomic Classification (Categorical): Each species was labeled with its Order and Family.
\item Diet Type (Categorical): Species were classified as herbivorous, carnivorous, omnivorous, or other specialized feeding types.
\item Activity Pattern (Categorical): Species were labeled as diurnal, nocturnal, crepuscular, or cathemeral based on their primary active periods.
\item Locomotion and Habitat Usage (True / False): We assigned Arboreal, Aquatic, Terrestrial, Fossorial, and Aerial to indicate primary movement and habitat types. Locomotion was further categorized into Quadrupedal or Bipedal.
\item Habitat Preferences (True / False): Each species was marked for its presence in Forest, Grassland, Desert, Wetland, Mountain, or Urban environments.
\item Climatic Distribution (True / False): Species were categorized based on their occurrence in Tropical, Subtropical, Temperate, Boreal, or Polar regions.
\item Social and Predatory Behavior (True / False): Herding indicates if the species lives in groups, while Predator specifies if it primarily hunts other animals.
\item Migratory Behavior (True / False): Species were marked as Migratory if they undergo seasonal long-distance movements.
\end{itemize}
\noindent\textbf{Dataset split.}
To ensure dataset quality, we do not separate images from the same iNaturalist entry into different dataset splits. Since a single entry can contain multiple images of the same subject, often taken from different angles, splitting them across training, validation, and test sets could lead to data leakage. Therefore, all images within the same submission are assigned to the same split. We divide the dataset into training, validation, and test sets using a 7:1:2 ratio.

\noindent\textbf{Frequency categorization.}
To gain additional insights during the evaluation of models trained on \datasetname, we categorized the dataset based on the training splits into three groups: frequent, intermediate, and rare categories. The taxonomy-based split was applied separately to footprints, feces, bones, eggs, and feathers. The top 20\% of categories were classified as frequent, the next 60\% as intermediate, and the bottom 20\% as rare.

\begin{table*}[t]
    \centering
    \setlength{\tabcolsep}{3pt} 
\scalebox{0.87}{
    \begin{tabular}{lccccc|ccccc|ccccc} \toprule
      \multirow{2}{*}{Model} &\multicolumn{5}{c}{Species}&\multicolumn{5}{c}{Family}&\multicolumn{5}{c}{Order}\\
      \cmidrule(lr){2-16}
       &Footprint&Feces&Egg&Bone&Feather&Footprint&Feces&Egg&Bone&Feather&Footprint&Feces&Egg&Bone&Feather\\ 
      \midrule
      \textit{All Categories}\\ 
       VGG-16~\cite{vgg} &28.8&29.6&45.2&14.7&56.7&45.6&46.6&61.1&31.0&66.1&62.1&65.1&81.2&54.2&78.7\\
       ResNet-50~\cite{resnet}  &23.7&29.4&41.1&18.3&59.7&41.7&48.6&59.9&33.6&70.1&58.9&64.8&80.8&53.0&81.8\\
       EFNet-B1~\cite{efficientnet}& 25.9&30.5&41.0&15.0&55.9&42.2&48.4&56.3&29.3&64.4&56.5&61.3&74.5&45.4&77.1\\
       ViT-B~\cite{vit}& 29.2&32.2&46.7&15.0&55.9&47.0&51.8&63.7&28.8&69.9&61.8&69.3&83.1&50.9&81.2\\
       Swin-B~\cite{swin}& 32.3&38.6&49.4&20.5&65.3&49.3&56.8&65.1&37.6&72.4&66.1&70.4&84.0&56.6&77.7\\
       \midrule
       \textit{Frequent Categories}\\ 
       ResNet-50~\cite{resnet}& 24.8&30.6&49.4&26.7&68.5&44.0&51.7&65.5&42.8&74.2&62.9&70.6&83.9&62.7&80.4\\ 
       Swin-B~\cite{swin}&33.7&40.3&58.5&25.6&65.3&52.4&60.6&70.0&47.8&72.4& 69.2&75.1&86.2&64.4&80.8\\
       \midrule
       \textit{Rare Categories}\\ 
        ResNet50~\cite{resnet}& 7.4&25.0&12.1&1.00&2.16&27.0&25.8&22.4&10.6&16.9&33.3&28.6&63.0&15.4&26.1\\ SwinB~\cite{swin}&14.2&28.1&14.1&4.91&2.52&40.4&22.7&26.3&15.4&22.1&44.4&31.0&66.7&11.5&30.4\\ 
      \bottomrule
    \end{tabular}
    }
  \caption{\textbf{Classification accuracy for all, frequent, and rare categories of animal species}. Throughout the species, family, and order categorization, Swin-B model tends to have higher accuracies on \datasetname.}
  \label{tb:classification_acc}
\end{table*}

\subsection{Statistics of \datasetname}
\label{sec:statistics}
\datasetname consists of 968 species, 200 families, and 65 orders. The dataset includes a total of 159,605 bounding boxes across five trace types:
\begin{itemize}
\item Footprints: 18,291 bounding boxes from 7,581 images, covering 117 species, 46 families, and 20 orders
\item Feces: 18,932 bounding boxes from 6,433 images, covering 101 species, 46 families, and 21 orders
\item Bones: 16,553 bounding boxes from 12,908 images, covering 269 species, 112 families, and 45 orders
\item Eggs: 29,434 bounding boxes from 9,394 images, covering 283 species, 67 families, and 20 orders
\item Feathers: 76,395 bounding boxes from 60,491 images, covering 555 species, 89 families, and 30 orders
\end{itemize}
The total number of bounding boxes matches that of the classification dataset, while the number of images aligns with the detection and segmentation datasets.
Additionally, we provide detailed trait annotations, including activity patterns, habitat types, climatic conditions, and behavioral characteristics, to facilitate animal tracking from multiple perspectives.
The species distribution across these trace sources is illustrated in Figure~\ref{fig:fig2}. We present the distribution of the top 100 species.

\section{Experiments}
We conduct experiments to validate and establish benchmarks on the proposed \datasetname. Here, we show the image classification, detection, instance segmentation, and traits classification results.

\subsection{Image Classification}
\label{sec:exp_cls}
\begin{figure*}
\centering
\includegraphics[width=1.0\linewidth]{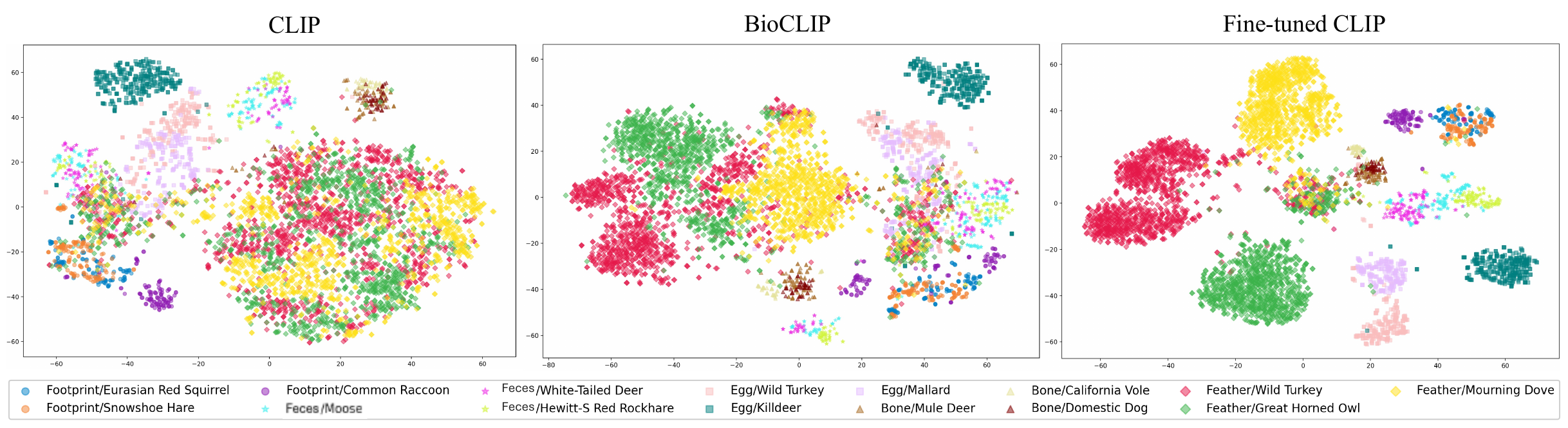}
\caption{\textbf{Visualization of t-SNE.} By using a labeled dataset specialized for observing indirect animal clues, the separability among categories has been improved. When visualized in the feature space, the categories are better distinguished.}
\vspace{-13pt}
\label{fig:tsne}
\end{figure*}

\paragraph{Settings.} As mentioned in Section~\ref{sec:data_annotation}, we cropped all animal traces using their assigned bounding boxes to minimize background effects as much as possible.

\paragraph{Baselines.}
We adopt five baseline models: VGGNet-16 (VGG-16)~\cite{vgg}, ResNet-50~\cite{resnet}, EfficientNet-B1 (EFNet-B1)~\cite{efficientnet}, Vision Transformer-Base (ViT-B)~\cite{vit}, and SwinTransformer-Base (Swin-B)~\cite{swin}. To allow convergence, we train the listed recognition models for 100 epochs on footprints, feces, bones, and eggs, and 50 epochs on feathers.

The implementation details are listed in the supplementary materials.

\paragraph{Metrics.}
We report the top-1 accuracy that measures the percentage of times the model's highest-probability prediction matches the correct label.

\paragraph{Results.}

We show the top-1 accuracy (\%) for different taxonomies and models in Table~\ref{tb:classification_acc}.
The results indicate that Swin-B~\cite{swin} tends to achieve the highest top-1 accuracy among the baseline models.
Among the trace types, feathers exhibit the highest accuracy across all taxonomies, despite having the highest number of species. Following feathers, eggs achieve the second-highest accuracy, which also corresponds to the second-largest number of species. As shown in Figure~\ref{fig:animal_images}, both feathers and eggs often feature distinctive colors and patterns, which may contribute to improved identification accuracy.
In contrast, bones exhibit lower accuracy. According to Figure~\ref{fig:animal_images}, bone appearance varies by body part, making it difficult to achieve high accuracy.
Additionally, we present the results for both frequent and rare categories of animal species, as described in Sec.~\ref{sec:data_annotation}. Swin-B also achieves higher scores on \datasetname.

\subsection{Feature Space Analysis}


\begin{table*}[t]
    \centering
    \setlength{\tabcolsep}{3pt} 
\scalebox{0.87}{
    \begin{tabular}{lccccc|ccccc|ccccc} \toprule
      \multirow{2}{*}{Model} &\multicolumn{5}{c}{Species}&\multicolumn{5}{c}{Family}&\multicolumn{5}{c}{Order}\\
      \cmidrule(lr){2-16}
       &Footprint&Feces&Egg&Bone&Feather&Footprint&Feces&Egg&Bone&Feather&Footprint&Feces&Egg&Bone&Feather\\ 
      \midrule
    \textit{All Categories}\\ YOLOv8~\cite{yolov8_ultralytics}&0.10&0.11&0.13&0.08&0.25&0.17&0.16&0.33&0.14&0.19&0.22&0.20&0.50&0.16&0.43\\
    YOLOv11~\cite{yolo11_ultralytics}&0.10&0.12&0.14&0.07&0.25&0.17&0.16&0.36&0.13&0.35&0.24&0.21&0.47&0.16&0.43\\
    Faster-RCNN~\cite{fasterrcnn}&0.04&0.06&0.12&0.07&0.08&0.07&0.09&0.17&0.0&0.12&0.08&0.13&0.26&0.10&0.22\\
    DINO~\cite{zhang2022dino}&0.08& 0.12 &0.20&0.07& 0.15 &0.12& 0.17 &0.32&0.17& 0.23&0.14& 0.22 &0.52&0.22&0.34\\
    RT-DETR~\cite{lv2023detrs}&0.10&0.17&0.04&0.01&0.17&0.21&0.25&0.42&0.17&0.40&0.31&0.28&0.57&0.21&0.50\\
           \midrule
       \textit{Frequent Categories}\\ 
       YOLOv11~\cite{yolo11_ultralytics}&0.17&0.18&0.38&0.11&0.56&0.29&0.26&0.63&0.29&0.70&0.33&0.40&0.73&0.44&0.81\\
        RT-DETR~\cite{lv2023detrs}&0.19&0.20&0.21&0.06&0.53&0.33&0.31&0.62&0.39&0.76&0.35&0.46&0.73&0.48&0.84\\ 
               \midrule
       \textit{Rare Categories}\\ 
       YOLOv11~\cite{yolo11_ultralytics}&0.04&0.05&0.04&0.003&0.14&0.07&0.08&0.07&0.05&0.05&0.10&0.02&0.26&0.007&0.005\\
        RT-DETR~\cite{lv2023detrs}&0.006&0.09&0.00&0.00&0.00&0.11&0.21&0.08&0.00&0.00&0.34&0.10&0.38&0.10&0.00\\ 
              \bottomrule
    \end{tabular}}
  \caption{\textbf{Detection results on footprints, feces, eggs, bones, and feathers datasets~(mAP@50-95).}}
  \vspace{-15pt}
  \label{tb:detection_map}
  
\end{table*}

\label{sec:exp_vis}
\paragraph{Baselines.}
We evaluate CLIP-based models and analyze their feature extraction ability by visualizing t-SNE. The models used in our evaluation include: a) ViT-B-32, the standard CLIP model~\cite{clip} pretrained by OpenAI.
b) BioCLIP~\cite{stevens2024bioclip}, a domain-specific CLIP model designed for biological image understanding.
c) Fine-tuned CLIP, a CLIP model trained on \datasetname using contrastive loss to improve performance on our specific dataset.

\paragraph{Metrics.}
For t-SNE visualization, we use a perplexity value of 30 and run 1000 iterations to ensure stable clustering. 

\paragraph{Results.}
Figure~\ref{fig:tsne} shows that our fine-tuned CLIP model achieves the best feature separation across categories. BioCLIP demonstrates better differentiation compared to the standard CLIP model, particularly in distinguishing between feathers and eggs. However, species-level feature separation remains challenging, especially for footprints and eggs.
This indicates that for animal trace identification, models need to be trained specifically on trace data, even if they have been pre-trained on a large dataset of animal images.

\subsection{Object Detection}
\label{sec:exp_det}
\noindent\textbf{Baselines.}
We adopt five baseline models: YOLOv8~\cite{yolov8_ultralytics}, YOLOv11~\cite{yolo11_ultralytics}, Faster-RCNN~\cite{fasterrcnn}, DINO~\cite{zhang2022dino}, RT-DETR~\cite{lv2023detrs}.
We train the models for 100 epochs for YOLOv8 and YOLOv11, and 50 epochs for DINO, Faster-RCNN, and RT-DETR.
The implementation details are found in the supplementary materials.

\noindent\textbf{Metrics.}
We report the mean average precision (mAP) over classes. 
mAP is computed by the mean of AP on different thresholds of intersection of interest (IoU), \ie, 0.5, 0.55, 0.60, ..., 0.95.

\noindent\textbf{Result.}
We present the bounding box mAP in Table~\ref{tb:detection_map}. Overall, RT-DETR achieves the best results across all categories. However, when examining rare categories, RT-DETR fails to detect them effectively, while its detection performance remains strong for frequent categories. The low mAP for rare categories contributes to the overall lower score across all categories.
Among the animal trace types, feathers achieved the best results in both major categories and overall. However, detecting minor categories remains challenging. This may be attributed to the wide category range of feathers.
\begin{figure}
\centering
\includegraphics[width=1.0\linewidth]{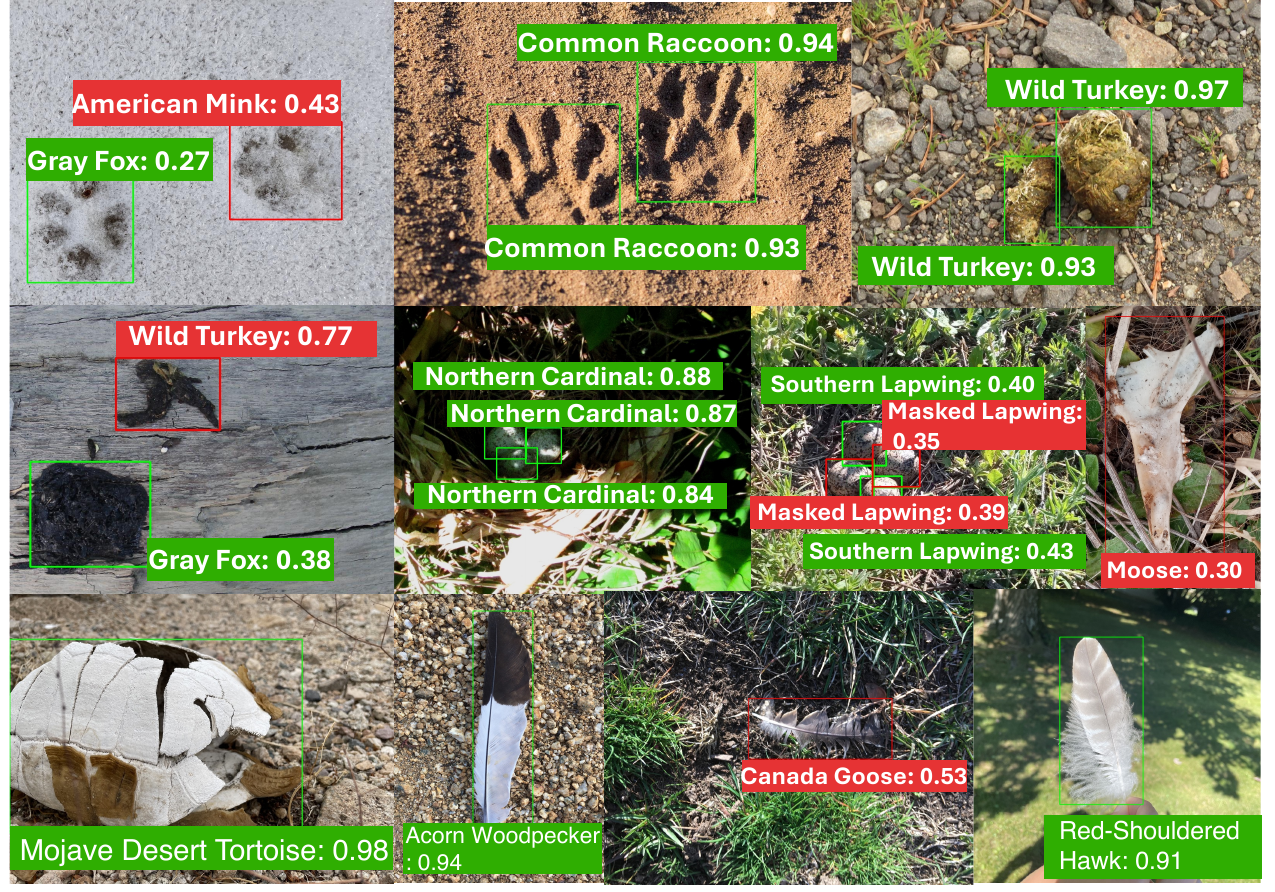}
\caption{\textbf{Visualization of YOLOv11 detection results.} The green bounding box denotes the correct detection, and the red bounding box denotes the wrong detection.}
\label{fig:detection}
\vspace{-15pt}
\end{figure}
In Figure~\ref{fig:detection}, we show the visualization of object detection on \datasetname using YOLOv11. The model correctly recognizes traces, including small ones, which are accurately identified at the species level. However, for faint footprints or traces in less visible areas, while they are recognized as traces, species misidentification is more common.
Since our dataset focuses on images in the wild, it presents a challenging benchmark.

\subsection{Instance Segmentation}
\label{sec:exp_seg}

\begin{table*}[t]
    \centering
    \begin{tabular}{lcccc|cccc|cccc} \toprule
      \multirow{2}{*}{Model} &\multicolumn{4}{c}{Species}&\multicolumn{4}{c}{Family}&\multicolumn{4}{c}{Order}\\
      \cmidrule(lr){2-13}
    &Feces&Egg&Bone&Feather&Feces&Egg&Bone&Feather&Feces&Egg&Bone&Feather\\ 
      \midrule
      \textit{All Categories}\\ YOLOv8~\cite{yolov8_ultralytics}&0.11&0.11&0.07&0.24&0.14&0.29&0.13&0.33&0.20&0.44&0.15&0.41\\
      YOLOv11~\cite{yolo11_ultralytics}&0.11&0.12&0.06&0.24&0.15&0.32&0.13&0.34&0.20&0.45&0.15&0.41\\
      Mask-RCNN~\cite{maskrcnn}&0.08&0.16&0.05&0.08&0.10&0.22&0.08&0.17&0.13&0.35&0.10&0.24\\
      MaskDINO~\cite{li2022mask}~&0.13&0.25&0.07&0.18&0.18&0.32&0.11&0.27&0.23&0.48&0.16&0.37\\
                 \midrule
       \textit{Frequent Categories}\\ YOLOv8~\cite{yolov8_ultralytics}&0.14&0.29&0.11&0.48&0.22&0.54&0.29&0.66&0.34&0.65&0.43&0.79\\ 
       YOLOv11~\cite{yolo11_ultralytics}&0.16&0.33&0.10&0.49&0.24&0.57&0.28&0.68&0.37&0.64&0.45&0.80\\ 
                  \midrule
       \textit{Rare Categories}\\ 
       YOLOv8~\cite{yolov8_ultralytics}&0.07&0.02&0.006&0.12&0.04&0.03&0.06&0.06&0.03&0.26&0.05&0.02\\ 
       YOLOv11~\cite{yolo11_ultralytics}&0.06&0.03&0.005&0.11&0.07&0.06&0.08&0.05&0.06&0.27&0.05&0.002\\ 
      \bottomrule
    \end{tabular}
  \caption{\textbf{Segmentation results on fecess, eggs, bones, and feathers datasets~(Mask mAP@50-95).}}
  \label{tb:segmentation_map}
\end{table*}

\begin{figure}
\centering
\includegraphics[width=1.0\linewidth]{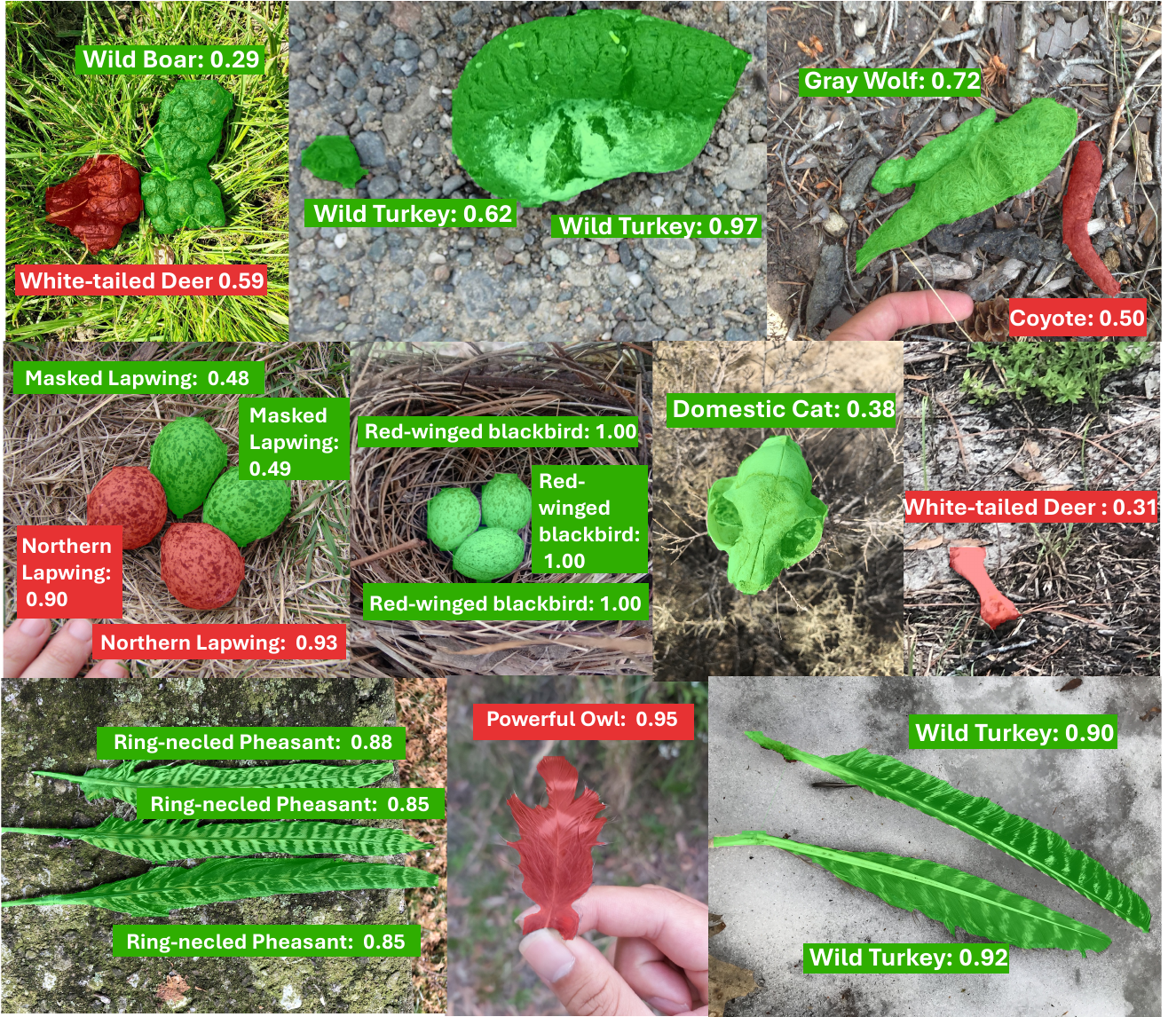}
\caption{\textbf{Visualization of YOLOv11 instance segmentation results.} The green mask denotes the correct mask, and the red mask denotes the wrong mask.}
\label{fig:segmentation}
\vspace{-13pt}
\end{figure}
\begin{table*}[t]
    \centering
    \begin{tabular}{lcccccccc} \toprule
       &Diet Type&Activity Pattern&Aquatic&Urban&Tropical&Polar&Herding&Predator\\ 
      \midrule
    Footprint&57.3~/~47.0&69.9~/~65.8&84.4~/~70.2&95.2~/~60.0&91.3~/~63.5&62.2~/~61.8 &84.2~/~73.3&80.7~/~94.3\\
       Feces&76.0~/~64.7&72.9~/~72.1&93.1~/~75.1&98.6~/~59.3&82.7~/~70.0&69.1~/~45.6&74.7~/~53.4&88.4~/~85.7\\
       Egg&68.7~/~48.2&95.4~/~73.0&88.4~/~83.4&76.8~/~75.2&90.6~/~69.7&93.8~/~75.4&82.5~/~81.5&96.9~/~73.4\\
       Bone&64.8~/~59.0&60.6~/~60.7&81.5~/~73.3&75.2~/~61.4&72.6~/~64.9&87.5~/~62.7&73.9~/~73.1&76.9~/~72.1\\
       Feather&81.3~/~80.0&89.7~/~63.7 &92.1~/~81.6&83.0~/~81.5&91.5~/~69.2&94.7~/~61.9&86.3~/~81.7&91.8~/~90.7\\

      \bottomrule
    \end{tabular}
  \caption{\textbf{Traits classification results on footprints, feces, eggs, bones, and feathers datasets~(Acc/F1 score).}}
  \label{tb:traits}
  \vspace{-10pt}
\end{table*}
\noindent\textbf{Baselines.}
We adopt four baseline models: YOLOv8~\cite{yolov8_ultralytics}, YOLOv11~\cite{yolo11_ultralytics}, Mask-RCNN~\cite{maskrcnn}, and MaskDINO~\cite{li2022mask}. For training, YOLOv8 and YOLOv11 run for 100 epochs, whereas Mask R-CNN and Mask DINO are trained for 50 epochs.
For implementation details, please refer to the supplementary materials.

\noindent\textbf{Metrics.}
We report the mean average precision (mAP) for segmentation across classes.
mAP is calculated as the mean of AP at different intersection over union (IoU) thresholds, \ie, 0.5, 0.55, 0.60, ..., 0.95.

\noindent\textbf{Result.} YOLOv8, YOLOv11 and MaskDINO achieve competitive results across three different taxonomic levels: species, family, and order. Overall, the trend is similar to the results from object detection as shown in Table~\ref{tb:segmentation_map}. Among animal traces, feathers tend to show high mAP despite having the highest number of species. This might be attributed to their distinct color appearance. Note that `footprint' does not contain labels for instance segmentation.
Therefore, we exclude it from this table.
In Figure~\ref{fig:segmentation}, we show the visual results of instance segmentation on \datasetname using YOLOv11.
The model tends to mislabel visually similar species (e.g., Masked Lapwing and Northern Lapwing, Gray Wolf and Coyote).

\subsection{Traits Classification}  
Next, we conduct experiments on trait attribute classification using animal traces. Among the 22 traits, we present classification results for Diet Type (six categories) and Activity Pattern (three categories), as well as Aquatic, Urban, Tropical, Polar, Herding, and Predator, which are binary (True/False) classifications.

\noindent\textbf{Baselines.} Here, we chose Swin Transformer~\cite{swin} for the baseline model, which achieved the best results in classification tasks.

\noindent\textbf{Metrics.} 
We report top-1 accuracy as a measure of overall classification performance and F1 score.

\noindent\textbf{Results.}
As shown in the Table~\ref{tb:traits}, feathers achieved the best results. For feces, the aquatic and predator traits showed high accuracy. This may be attributed to differences in feces characteristics between aquatic and terrestrial animals and the fact that higher hierarchical animals, such as predators, tend to have larger feces. Inferring an animal's traits from its traces may depend on the type of trace, with certain traces being more suitable for identifying specific traits.

\section{Conclusion}
We introduced \datasetname, a dataset for identifying animal species based on their traces, including footprints, feces, eggs, bones, and feathers. Our comprehensive annotations for classification, detection, and pixel-level segmentation, along with 22 traits and taxonomy annotations, bridge the gap between animal tracking and computer vision.
Our experiments show that there is still room for improvement, especially in rare category detection and segmentation.
As animal tracking plays a crucial role in understanding animal movements in a given region and is a non-invasive approach, we hope our dataset contributes to advancing research in animal tracking and wildlife conservation.
\section{Acknowledgment}
This research was supported by the AIST policy-based budget project “R\&D on Generative AI Foundation Models for the Physical Domain” and JST FOREST Grant Number JPMJFR206F. We used ABCI 3.0 provided by AIST and AIST Solutions.

{
    \small
    \bibliographystyle{ieeenat_fullname}
    \bibliography{main}
}

\clearpage
\appendix
\section{Implementation Details}
\paragraph{Classification.}
We train VGG-16~\cite{vgg}, ResNet-50~\cite{resnet}, ViT-B~\cite{vit}, and Swin-B~\cite{swin} with input images resized to $224^2$ pixels, while Eff-b1~\cite{efficientnet} uses $240^2$ pixels.  
Data augmentation includes random adjustments to brightness, contrast, saturation, hue, and compression rate, as well as vertical/horizontal flipping and rotation.  
For training, we use a batch size of 128 and the SGD optimizer with a learning rate of $1 \times 10^{-4}$.  
Models are trained for 100 epochs, except for the feather category, which is trained for 50 epochs. This setup ensures that training losses converge.  

\paragraph{Detection.}
We train YOLOv8~\cite{yolov8_ultralytics}, YOLOv11~\cite{yolo11_ultralytics}, Faster-RCNN~\cite{fasterrcnn}, DINO~\cite{zhang2022dino}, and RT-DETR~\cite{lv2023detrs} with input images resized to $512^2$ pixels.
We train the models for 100 epochs for
YOLOv8 and YOLOv11, and 50 epochs for DINO, Faster-RCNN, and RT-DETR.
For the implementation, we follow YOLOv8 and YOLOv11, and RT-DETR for Ultralytics~\cite{ultralytics}, Faster-RCNN for Detectron2~\cite{wu2019detectron2}, and DINO for detrex~\cite{ren2023detrex}. 

For Faster R-CNN, we used the Adam optimizer with a learning rate of 1e-4. The ROI head batch size per image was set to 256, and the batch size per iteration was 4.
For YOLOv8 and YOLOv11, we set the initial learning rate (lr0) to 0.01, with the final learning rate (lrf) defined as 0.01 × lr0. The batch size was 8.
For DINO, we used the DINO-R50-4scale model, which incorporates a ResNet-50 backbone and extracts features from four different resolution levels to enhance multi-scale object detection. The learning rate was set to 1e-4, with a batch size of 16.
For RT-DETR, we used the RT-DETR-L model with the AdamW optimizer and a learning rate of 0.001.

\paragraph{Instance Segmentation.}
We trained YOLOv8~\cite{yolov8_ultralytics}, YOLOv11~\cite{yolo11_ultralytics} for 100 epochs, and Mask-RCNN~\cite{maskrcnn}, and MaskDINO~\cite{li2022mask} for 50 epochs with input images resized to $512^2$ pixels.

For YOLOv8 and YOLOv11, we set the initial learning rate (lr0) to 0.01, with the final learning rate (lrf) defined as 0.01 × lr0. The batch size was 8.
For Mask R-CNN, the batch size per iteration was 4 images, with a learning rate of 1e-4. Additionally, the batch size per image for the Region of Interest (ROI) heads was set to 256.
For MaskDINO, we used the ResNet-50 backbone, setting the learning rate to 1e-4 and using a batch size of 16.

\section{Removed Images}
Here, we describe the types of images that are unsuitable for our \datasetname and were manually removed.

\begin{itemize}
\item Images with overlaid text: Citizen scientists sometimes write labels directly on the images.
\item Images containing animals: In many citizen scientists' posts, images include evidence of the animal itself to support the provided labels.
\item Distant subjects: Some posts include images where the subject appears too far away.
\item Images containing human faces: Although these images are licensed under Creative Commons, we exclude them from our dataset to protect privacy.
\end{itemize}

\begin{figure*}
\centering
\includegraphics[width=0.9\linewidth]{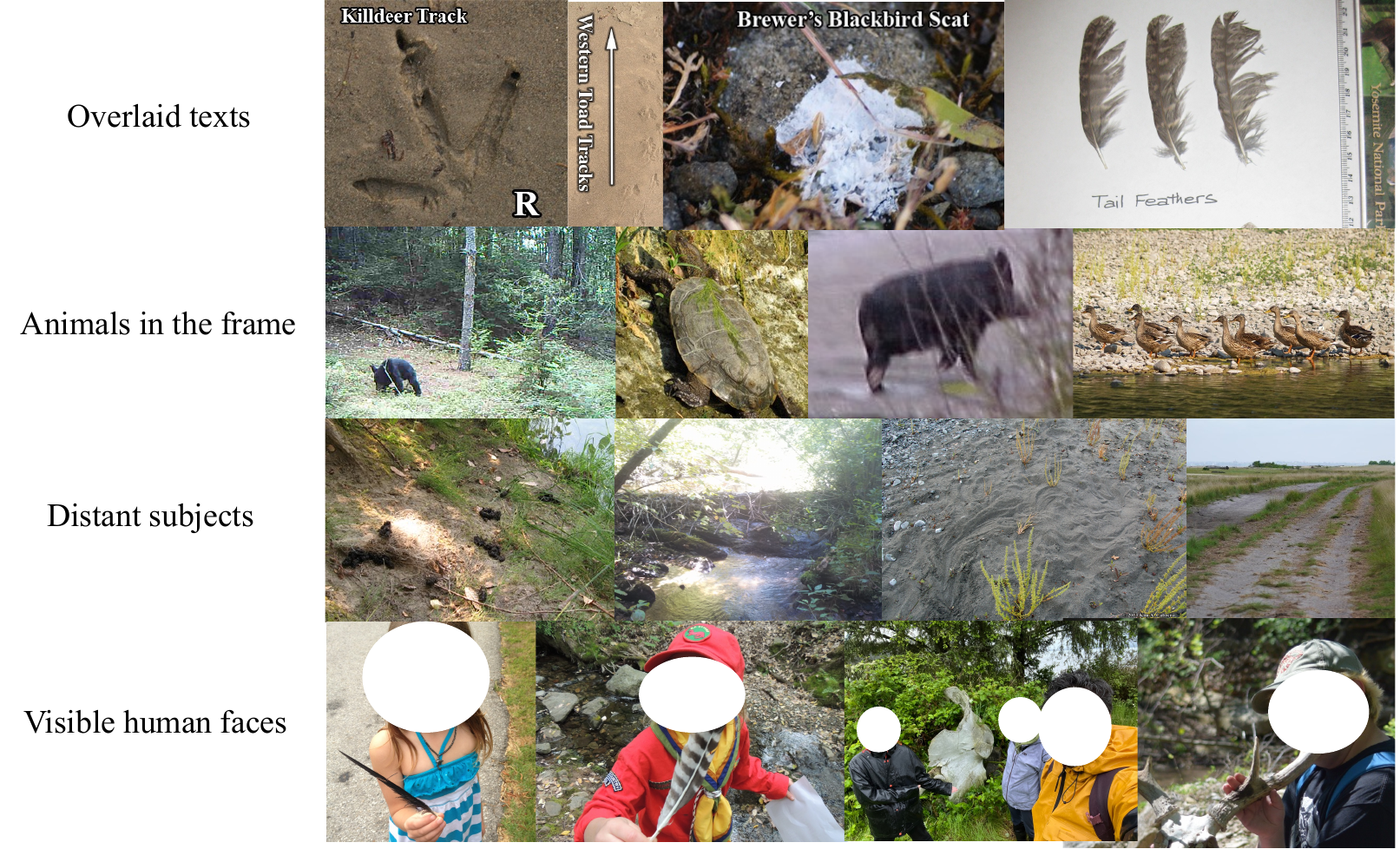}
\caption{\textbf{Examples of removed images from our dataset.}}
\label{fig:removed_images}
\end{figure*}

Figure~\ref{fig:removed_images} shows examples of removed images.

\section{Species Distribution}
\begin{figure*}
\centering
\includegraphics[width=1.0\linewidth]{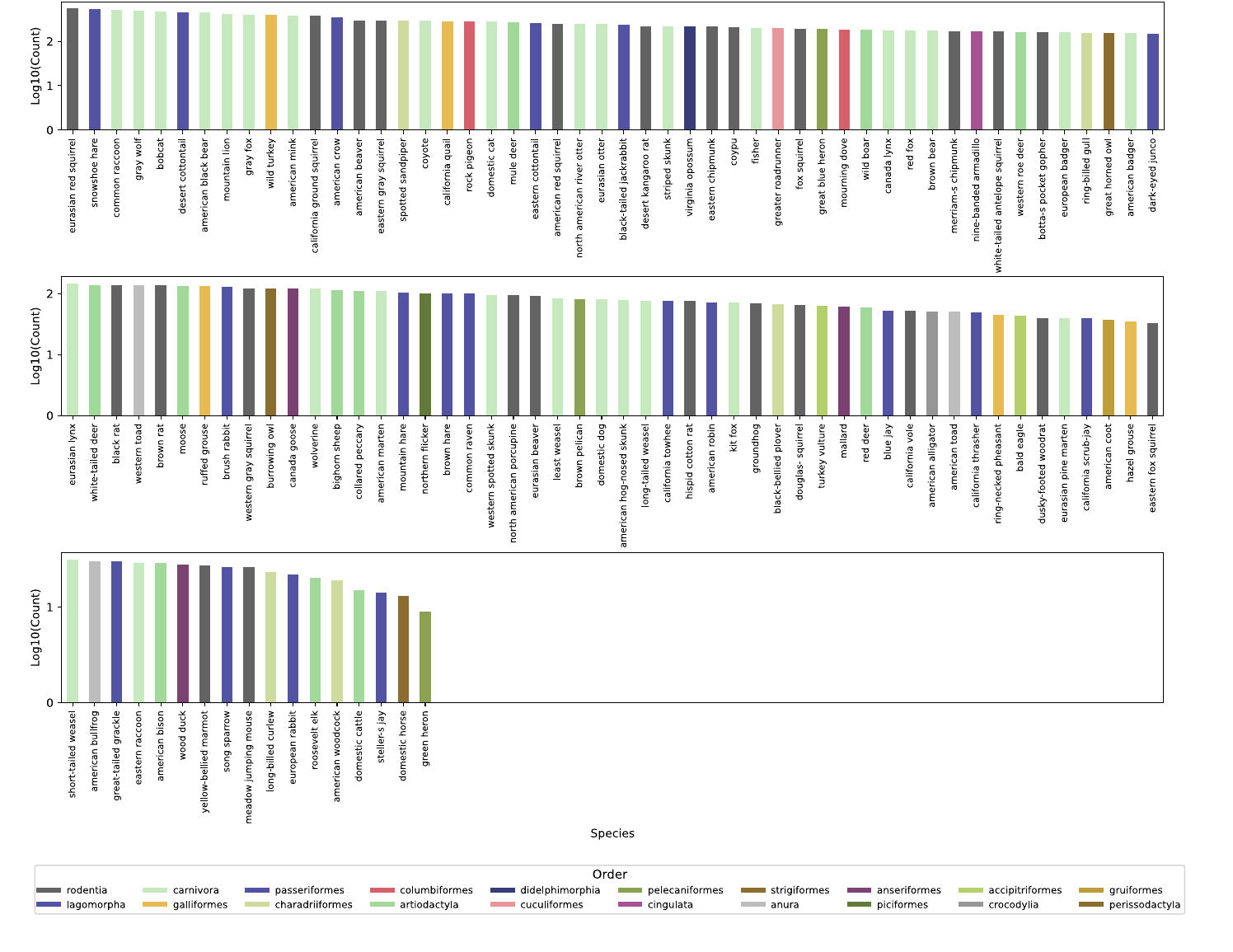}
\caption{\textbf{Species distributions of footprints.}}
\label{fig:foot}
\end{figure*}

\begin{figure*}
\centering
\includegraphics[width=1.0\linewidth]{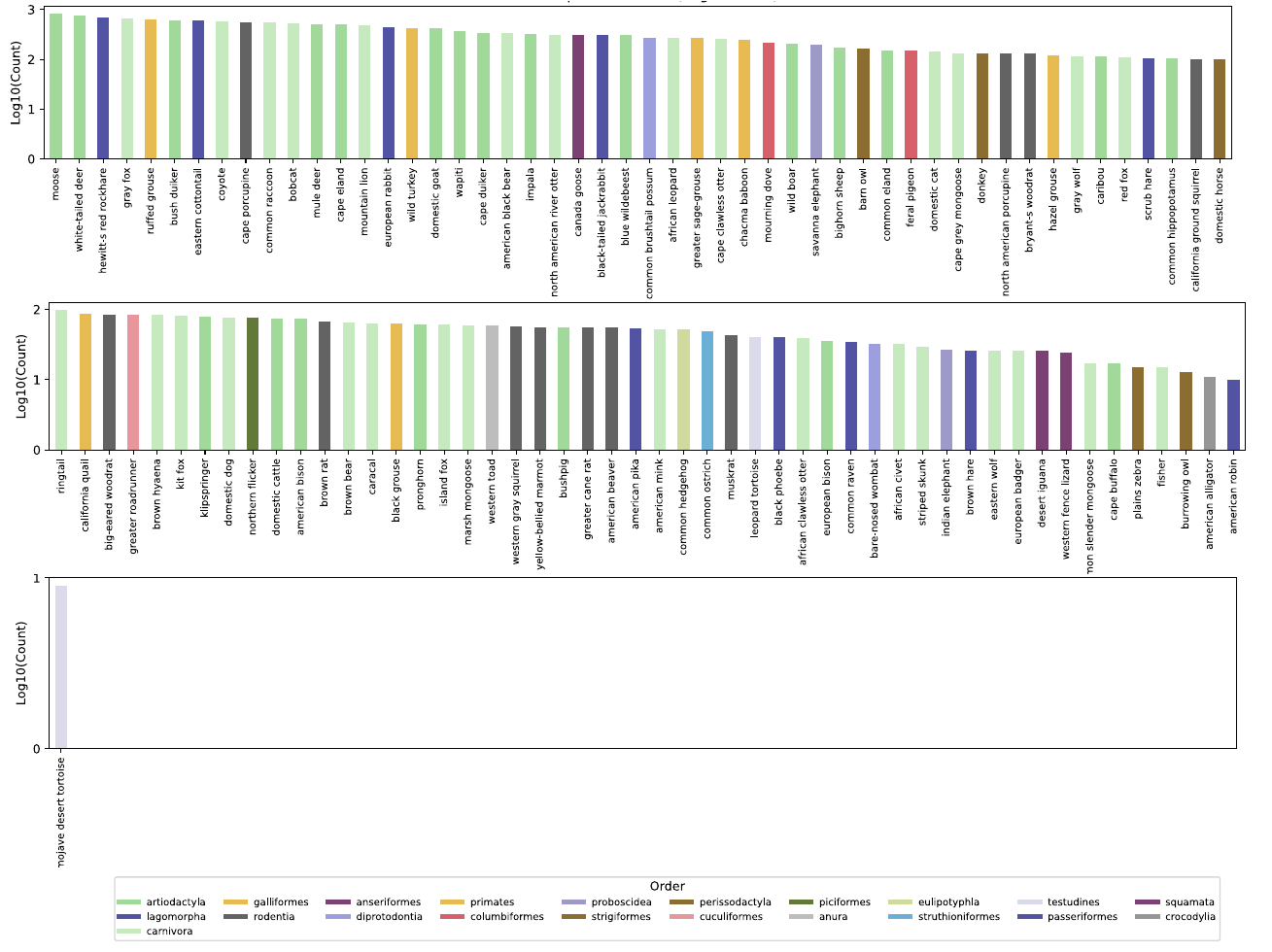}
\caption{\textbf{Species distributions of feces.}}
\label{fig:feces}
\end{figure*}

\begin{figure*}
\centering
\includegraphics[width=1.0\linewidth]{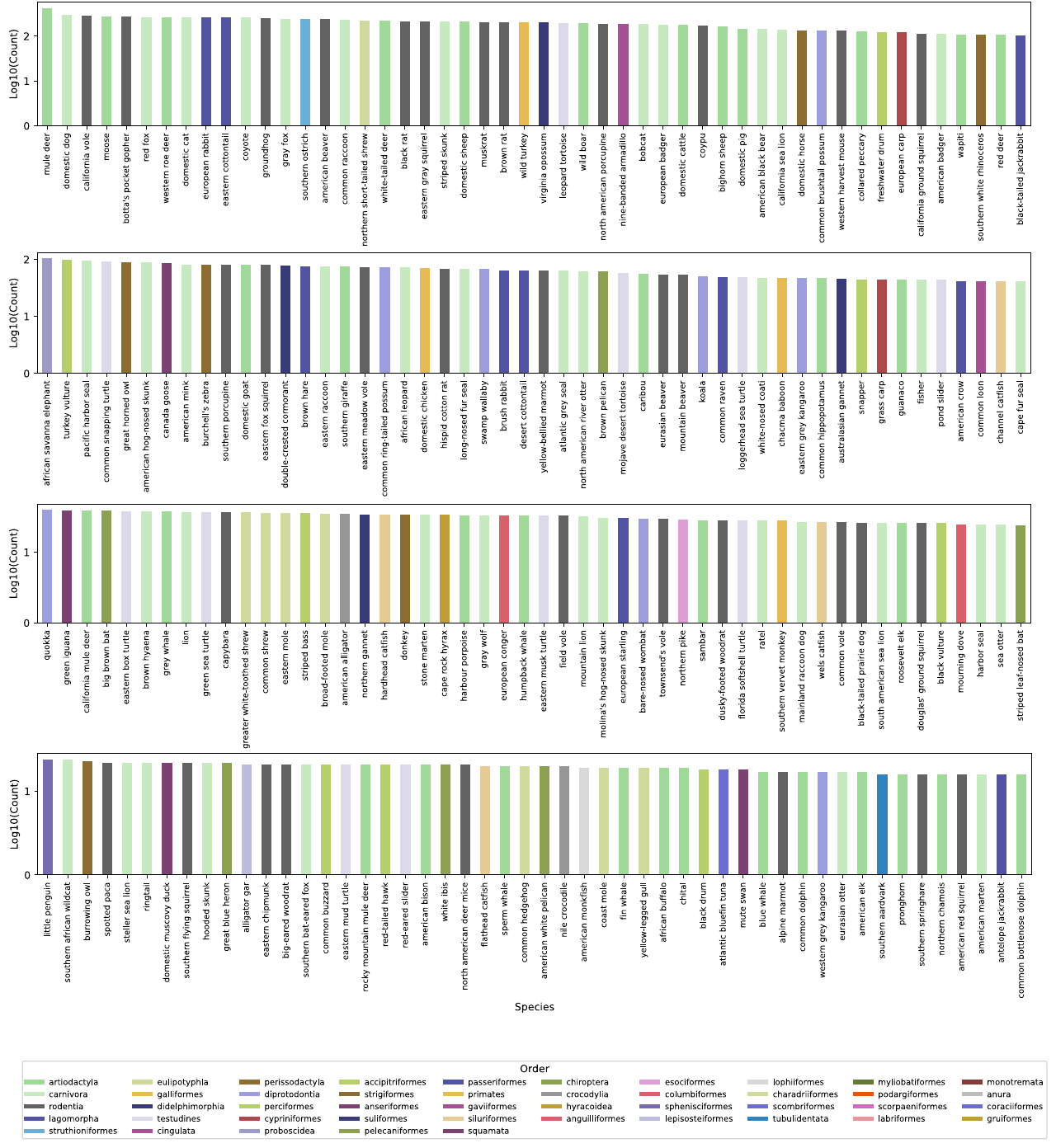}
\caption{\textbf{Species distributions of bones (A).}}
\label{fig:bonea}
\end{figure*}

\begin{figure*}
\centering
\includegraphics[width=1.0\linewidth]{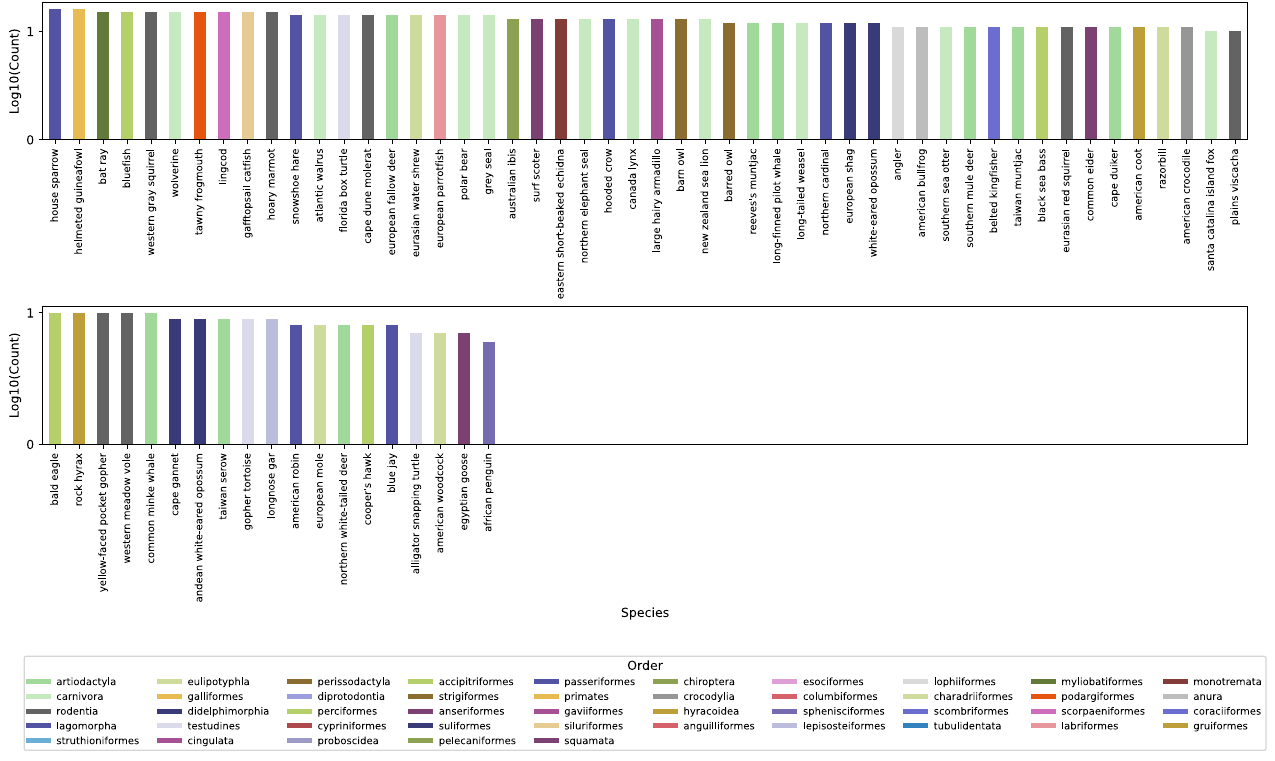}
\caption{\textbf{Species distributions of bones (B).}}
\label{fig:boneb}
\end{figure*}

\begin{figure*}
\centering
\includegraphics[width=1.0\linewidth]{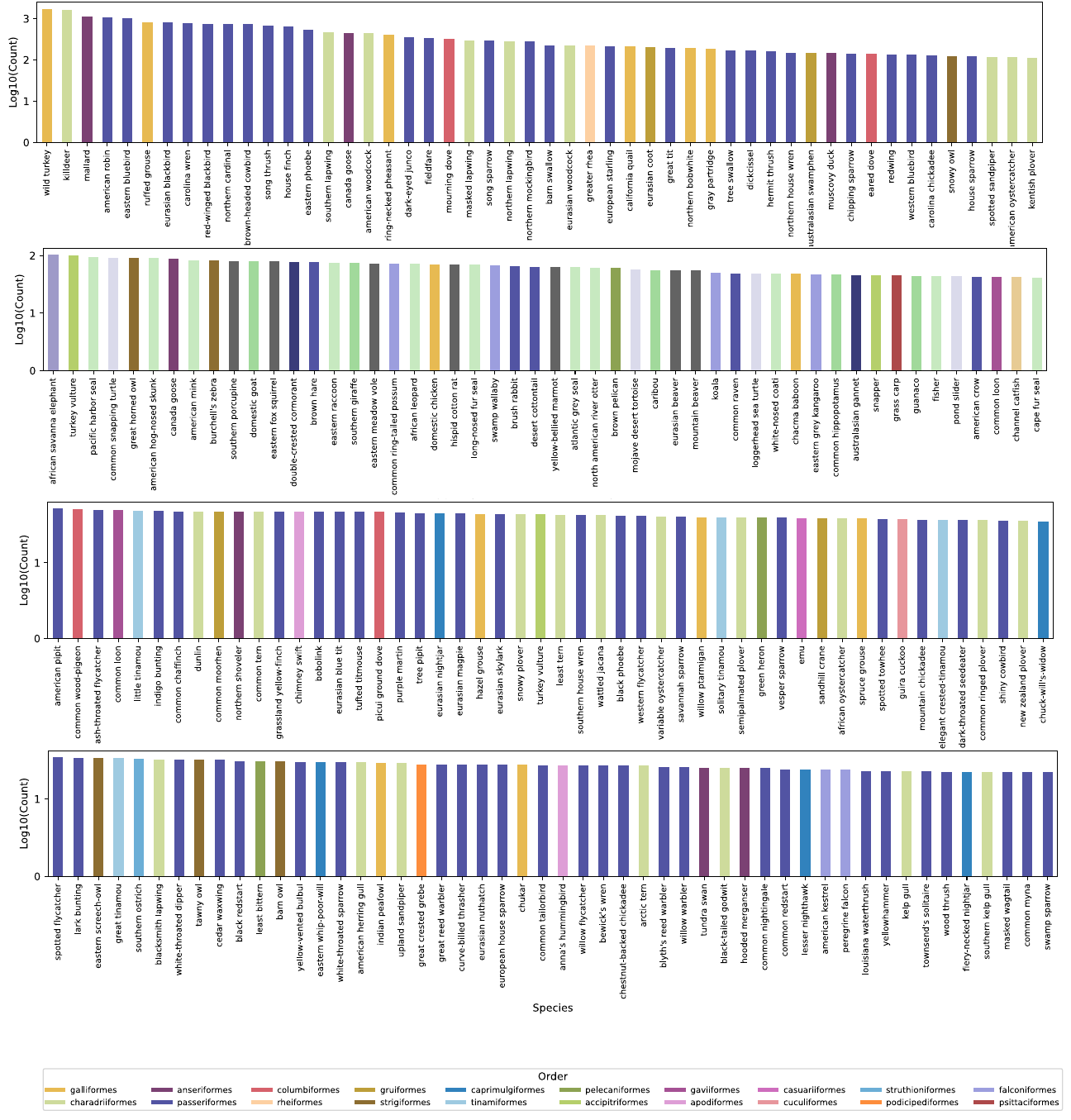}
\caption{\textbf{Species distributions of eggs (A).}}
\label{fig:egga}
\end{figure*}

\begin{figure*}
\centering
\includegraphics[width=1.0\linewidth]{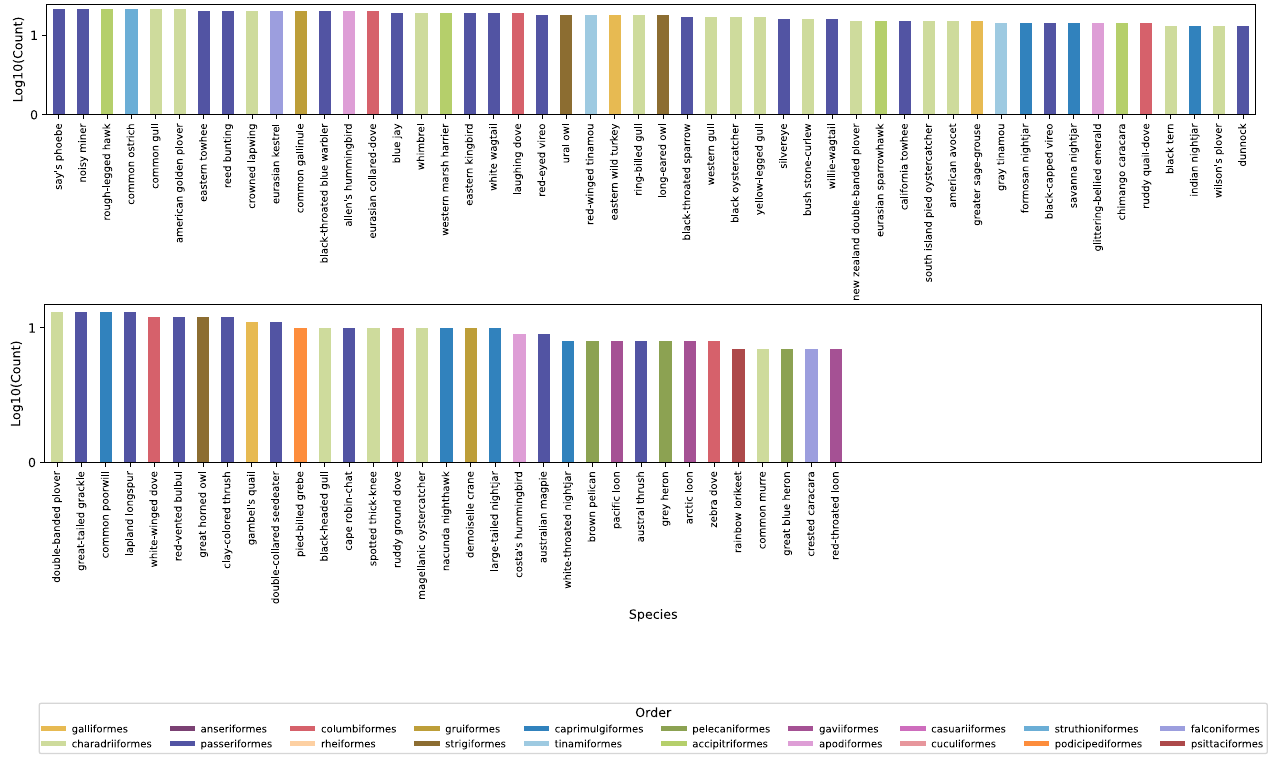}
\caption{\textbf{Species distributions of eggs (B).}}
\label{fig:eggb}
\end{figure*}

\begin{figure*}
\centering
\includegraphics[width=1.0\linewidth]{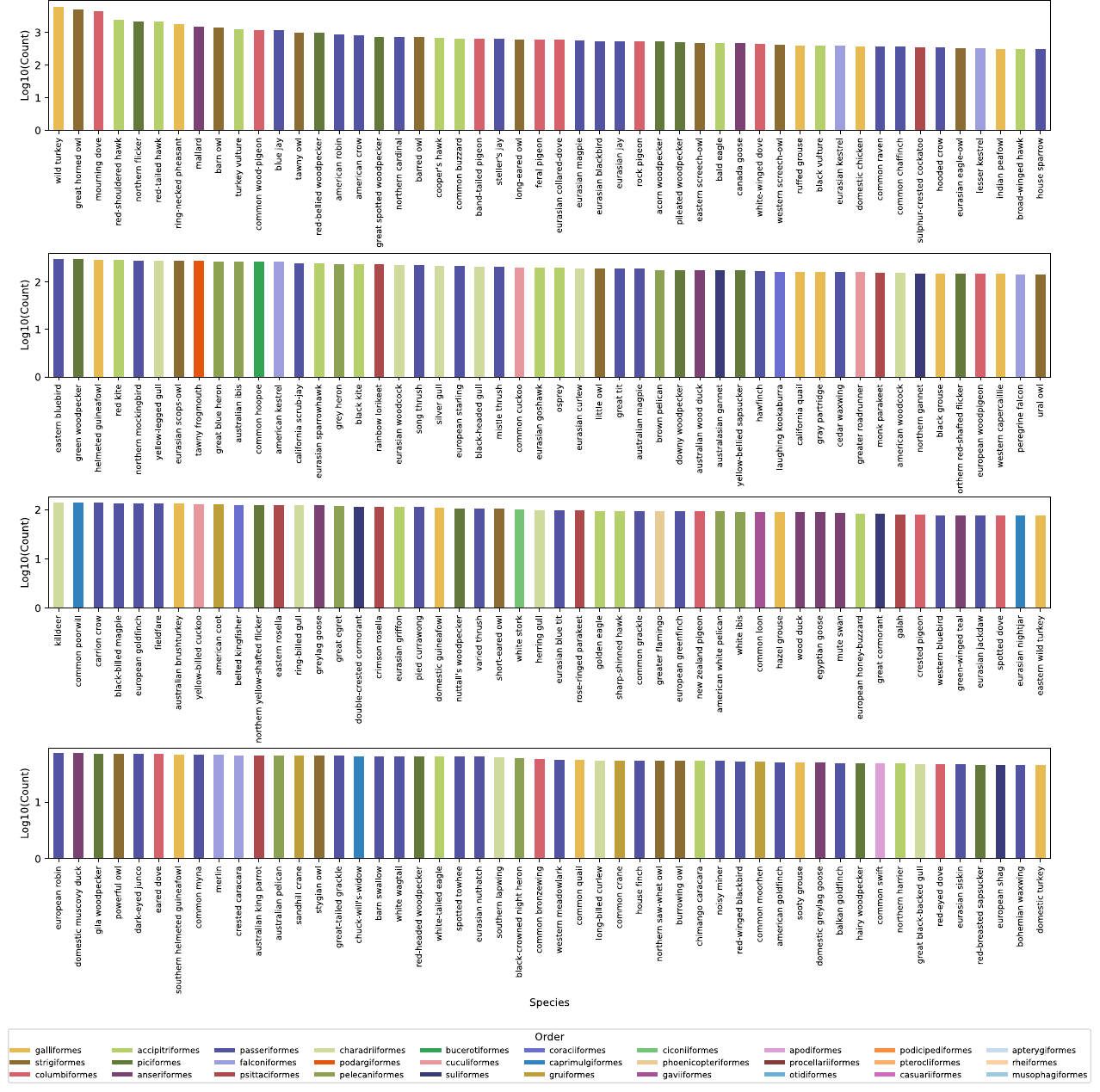}
\caption{\textbf{Species distributions of feathers (A).}}
\label{fig:feathera}
\end{figure*}

\begin{figure*}
\centering
\includegraphics[width=1.0\linewidth]{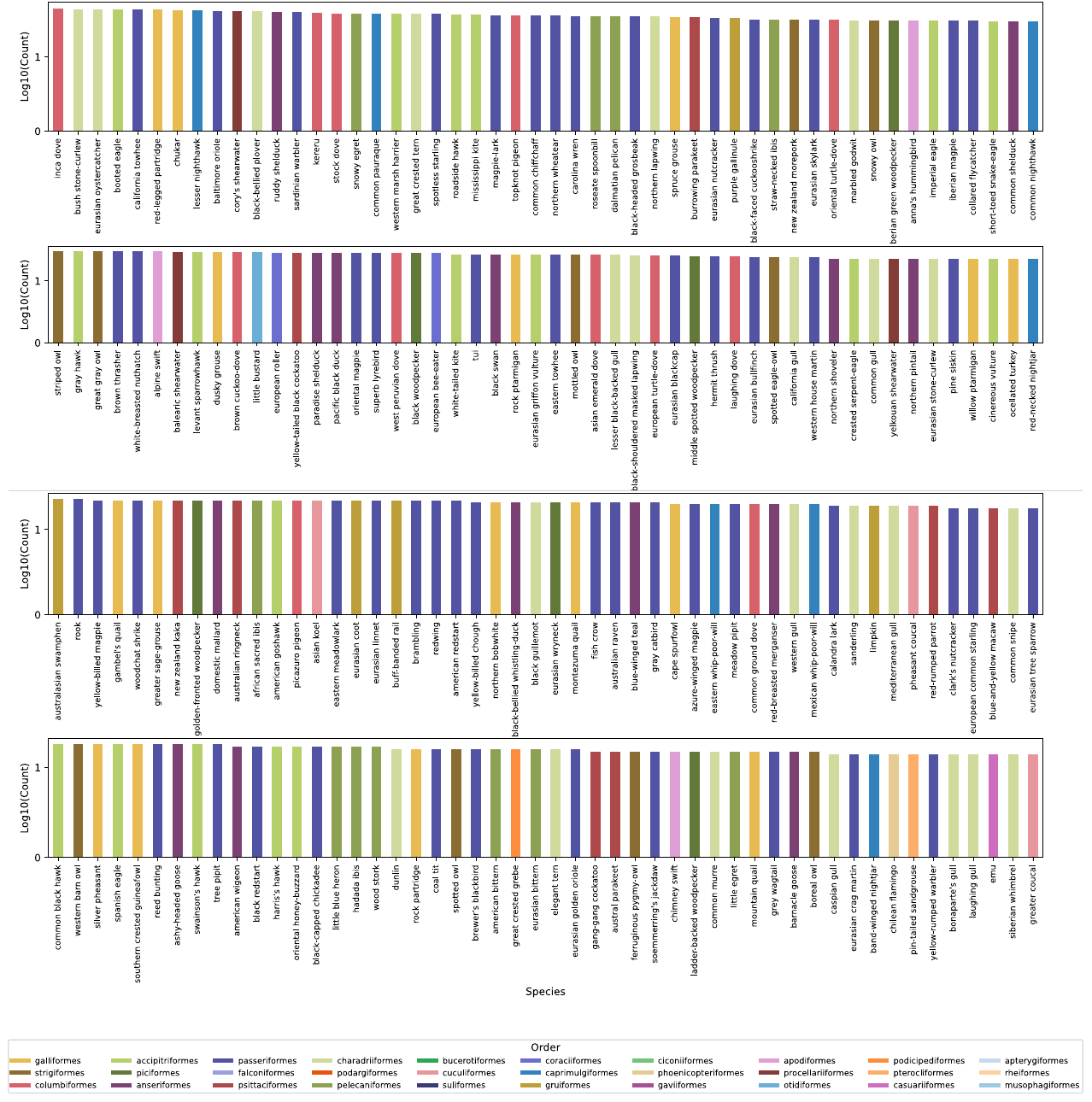}
\caption{\textbf{Species distributions of feathers (B).}}
\label{fig:featherb}
\end{figure*}

\begin{figure*}
\centering
\includegraphics[width=1.0\linewidth]{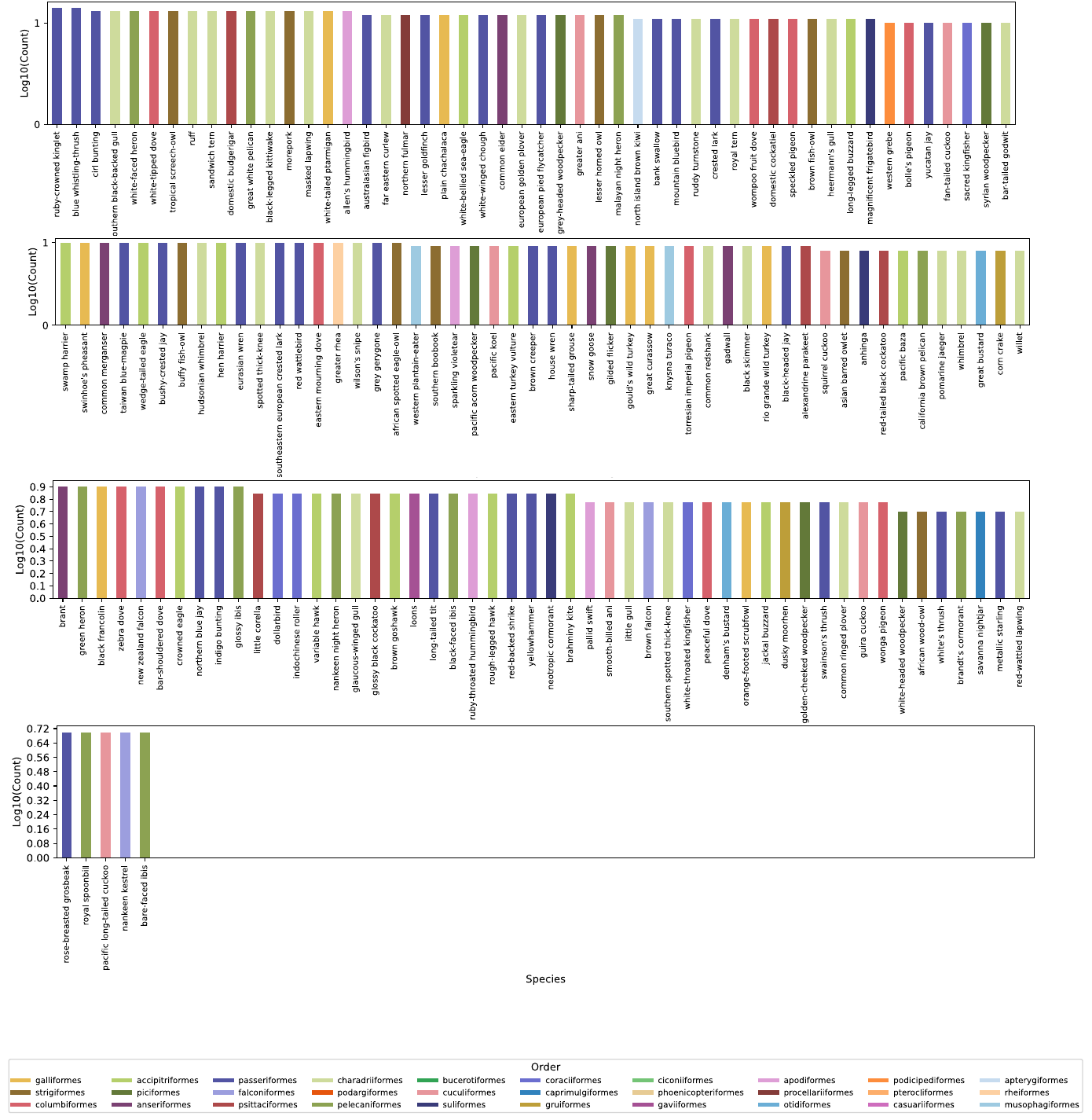}
\caption{\textbf{Species distributions of feathers (C).}}
\label{fig:featherc}
\end{figure*}

In this study, we collect five types of traces:
\begin{itemize}
\item Footprints: 117 species, 46 families, and 20 orders
\item Feces: 101 species, 46 families, and 21 orders
\item Bones: 269 species, 112 families, and 45 orders
\item Eggs: 283 species, 67 families, and 20 orders
\item Feathers: 555 species, 89 families, and 30 orders
\end{itemize}
We visualize the species distribution categorized by order:
\begin{itemize}
\item Footprints: Figure~\ref{fig:foot}
\item Feces: Figure~\ref{fig:feces}
\item Bones: Figure~\ref{fig:bonea}, Figure~\ref{fig:boneb}
\item Eggs: Figure~\ref{fig:egga}, Figure~\ref{fig:eggb}
\item Feathers: Figure~\ref{fig:feathera}, Figure~\ref{fig:featherb}, Figure~\ref{fig:featherc}
\end{itemize}
Additional trait details and taxonomy annotations will be made publicly available.


\section{Additional Experiments}
\paragraph{Other Metrics.} 
For reference, we also report the Top-5 accuracy and balanced accuracy in Table~\ref{tb:additional}. The model used is Swin-B with species-level classification.
\begin{table}[t]
    \centering
    \begin{tabular}{lccccc} \toprule
    &Footprint&Feces&Egg&Bone&Feather\\
          \midrule
 Top5&64.11&68.62&73.63&41.46&82.43\\
 Balanced&28.92&35.41&32.25&15.54&26.55\\
      \bottomrule
    \end{tabular}
\caption{\textbf{Species classification results on the footprints, feces, eggs, bones, and feathers datasets (Top-5 accuracy and balanced accuracy)}. The model used is Swin-B.}
  \label{tb:additional}
  \vspace{-10pt}
\end{table}

\paragraph{CLIP Training.} 
We conducted BioCLIP~\cite{stevens2024bioclip}-style training. We augmented the text prompts during CLIP training to include both a simple description and a taxonomic context. Specifically, we used the following format: "A photo of a \{species\}." and \"A photo of a \{species\}, which belongs to \{genus\}, \{family\}, \{order\}, \{class\}."
This training strategy led to a noticeable performance gain, improving the Top-1 accuracy from 17.6\% to 21.7\% on footprint species classification. 

\end{document}